\documentclass[sigconf]{acmart}

\usepackage{tcolorbox}
\usepackage{enumitem}
\usepackage{xcolor}
\usepackage{multirow}
\usepackage{comment}
\definecolor{awesome}{RGB}{70,130,180}

\AtBeginDocument{%
  }

\copyrightyear{2026}
\acmYear{2026}
\setcopyright{cc}
\setcctype{by}
\acmConference[MSR '26]{23rd International Conference on Mining Software Repositories}{April 13--14, 2026}{Rio de Janeiro, Brazil}
\acmBooktitle{23rd International Conference on Mining Software Repositories (MSR '26), April 13--14, 2026, Rio de Janeiro, Brazil}
\acmPrice{}
\acmDOI{10.1145/3793302.3793593}
\acmISBN{979-8-4007-2474-9/2026/04}

\newcommand{\RqOne}{\textbf{RQ1:} \emph{How do PR description characteristics differ across AI coding agents?}}
\newcommand{\RqTwo}{\textbf{RQ2:} \emph{How do human reviewers respond to PRs submitted by different AI coding agents?}}
\newcommand{\RqTwoone}{\textbf{RQ2-1:} \emph{How do reviewer response differ across AI coding agents?}}
\newcommand{\RqTwotwo}{\textbf{RQ2-2:} \emph{How do review outcomes differ across AI coding agents?}}

\begin{document}

\title{How AI Coding Agents Communicate: A Study of Pull Request Description Characteristics and Human Review Responses}

\author{
  Kan Watanabe\textsuperscript{$\dagger$},
  Rikuto Tsuchida \textsuperscript{$\dagger$},
  Takahiro Monno\textsuperscript{$\dagger$},
  Bin Huang\textsuperscript{$\dagger$},
  Kazuma Yamasaki\textsuperscript{$\dagger$},
  Youmei Fan\textsuperscript{$\dagger$},
  Kazumasa Shimari\textsuperscript{$\dagger$},
  Kenichi Matsumoto\textsuperscript{$\dagger$}
}

\affiliation{%
  \institution{\textsuperscript{$\dagger$}Nara Institute of Science and Technology, Japan}
  \country{}
}


\email{{watanabe.kan.wi6, tsuchida.rikuto.tq5, monno.takahiro.mv3, huang.bin.gv2,yamasaki.kazuma.yj9}@naist.ac.jp}
\email{{fan.youmei.fs2, k.shimari, matumoto}@is.naist.jp}

\renewcommand{\shortauthors}{Watanabe, et al.}

\begin{abstract}
The rapid adoption of large language models has led to the emergence of AI coding agents that autonomously create pull requests on GitHub. However, how these agents differ in their pull request description characteristics, and how human reviewers respond to them, remains underexplored. In this study, we conduct an empirical analysis of pull requests created by five AI coding agents using the AIDev dataset.
We analyze agent differences in pull request description characteristics, including structural features, and examine human reviewer response in terms of review activity, response timing, sentiment, and merge outcomes. 
We find that AI coding agents exhibit distinct PR description styles, which are associated with differences in reviewer engagement, response time, and merge outcomes.
We observe notable variation across agents in both reviewer interaction metrics and merge rates. These findings highlight the role of pull request presentation and reviewer interaction dynamics in human–AI collaborative software development.

\end{abstract}


\maketitle

\section{Introduction}
The advent of Large Language Models (LLMs) has catalyzed the widespread adoption of AI coding agents in software development on GitHub.
Unlike traditional code completion assistants, these agents autonomously modify code and submit complete pull requests (PRs) \cite{yang2024sweagentagentcomputerinterfacesenable, tufano2024autodevautomatedaidrivendevelopment}.
Li et al. characterize this paradigm shift as ``Software Engineering 3.0 (SE 3.0),'' identifying tools such as Github Copilot \cite{github_copilot} and Cursor \cite{Cursor} as partners that actively participate in the development process \cite{hassan2024ainativesoftwareengineeringse, li2025aidev}.

AI tools play a versatile role in enhancing coding productivity \cite{peng2023impactaideveloperproductivity,he2025doesaiassistedcodingdeliver,Mozannar2024}, streamlining workflows \cite{cihan2024automatedcodereviewpractice}, and detecting bugs \cite{shah2025evolutionprogrammerstrustgenerative}. Consequently, many developers express high satisfaction with these tools \cite{bakal2025experiencegithubcopilotdeveloper,vaillant2024developersperceptionsimpactchatgpt}, a sentiment largely stemming from the perceived quality and utility of the AI's output \cite{vaillant2024developersperceptionsimpactchatgpt}.

However, despite these advancements, significant barriers remain regarding the integration of generated code into real-world repositories.
The challenge that many PRs created by AI tools and agents are not accepted \cite{DBLP:journals/corr/abs-2110-15447, li2025aidev} may not be explainable solely by the functional correctness of the code.
In human-centric software development, even functionally correct code is frequently rejected due to insufficient explanation \cite{DBLP:journals/corr/abs-2504-18407}.
This challenge is amplified in human-AI collaboration, where programmers face a significant cognitive burden to verify that generated code aligns with their intent\cite{barke2023grounded}, suggesting that the communication of changes may be a potential bottleneck in human–AI collaborative development.
Therefore, accelerating the adoption of SE 3.0 requires AI coding agents to possess the capability to appropriately articulate and convey their changes. Furthermore, it is essential to ensure high satisfaction not only for the developers using the tools but also for the reviewers evaluating the contributions.

In this study, we focus on the characteristics of PR descriptions generated by AI coding agents and the corresponding human responses.
Specifically, we address the following research questions:

\noindent
-- \RqOne 

In contrast to functional performance differences such as merge rates revealed by prior work \cite{li2025aidev}, RQ1 focuses on ``PR description characteristics'' that directly influence reviewer cognitive load.

\noindent
-- \RqTwo

While developer satisfaction with AI tools is high, reviewers may treat PRs more critically due to the nature of contributions by AI coding agents or the cognitive burden imposed by their descriptions.
We analyze review interactions to determine whether humans respond differently to different AI teammates.
To answer this question, we investigate the following two specific aspects:

\noindent
-- \RqTwoone

\noindent
-- \RqTwotwo

\textbf{Replication Packages:} To facilitate replication and further
studies, we provide the data used in our replication package.\footnote{\url{https://doi.org/10.5281/zenodo.18031372}} 

\textbf{Contributions}
\begin{itemize}
    \item We provide the first large-scale empirical comparison of pull request description characteristics across multiple autonomous AI coding agents.
    \item We analyze how human reviewers interact with and respond to PRs generated by different AI coding agents in terms of engagement, timing, sentiment, and outcomes.
    \item We derive implications for designing AI coding agents that can better integrate into human-centric code review workflows.
\end{itemize}

\section{Approach}

\subsection{PR Description Characteristics (RQ1)} 
To address RQ1, we analyzed a total of 33,596 PRs generated by five AI coding agents (GitHub Copilot, OpenAI Codex, Claude Code, Devin, and Cursor), sourced from the \textit{pull\_request} table of the AIDev dataset.
Table~\ref{tab:RQ1features} details the metrics definitions used to quantify the PR description characteristics of each agent.

First, ``Work Style'' represents the coding style of each agent. These metrics are obtained by aggregating data from the \textit{pr\_commit\_\allowbreak details} and \textit{pr\_commits} tables.
\textit{files\_changed} is the count of unique filenames modified within a PR.
\textit{additions} and \textit{deletions} represent lines of code added and removed, respectively.
\textit{commits} is the number of commits included in the PR, derived from the commit log.

Next, ``Description Style'' indicates the PR description characteristics of each agent. These metrics are extracted from the PR body text using regular expressions.
To account for variations in description length, all metrics except \textit{char} are normalized as density per 1,000 characters.
\textit{char} is the raw character count of the description, indicating overall verbosity.
\textit{header} is the density of Markdown header tags (lines starting with ``\#'' followed by a space).
\textit{list} is the density of list items (lines starting with a bullet point, number, or plus sign followed by a space).
\textit{code} is the density of code blocks, calculated by counting occurrences of triple backticks (\`{}\`{}\`{}) and dividing the count by two.
\textit{emoji} is the density of emojis identified by specific Unicode ranges (e.g., U+1F600--U+1F64F).
\textit{politeness} is the density of case-insensitive matches against 15 polite phrases (please, thank, appreciate, grateful, sorry, apolog(y|ize|ies), kindly, could you, would you, if possible, feel free to, happy to, let me know, hope (this|that) helps, regards).

Finally, ``PR Compliance'' assesses adherence to PR title conventions.
\textit{conventional\_commit} is a binary variable indicating whether the PR title starts with one of the standard types (feat, fix, docs, chore, refactor, style, test, build, ci, perf, revert), optionally followed by a scope in parentheses, and ends with a colon and a space.

\begin{table}[tb]
\centering
\caption{features extracted for PR Description Characteristics}
\label{tab:RQ1features}
\resizebox{\columnwidth}{!}{%
\begin{tabular}{cll}
\toprule
Category                                                                     & \multicolumn{1}{c}{Metrics} & \multicolumn{1}{c}{Description} \\ 
\midrule
\multirow{4}{*}{\begin{tabular}[c]{@{}c@{}}Work\\ Style\end{tabular}}  & files\_changed              & Number of changed files         \\
                                                                             & additions                   & Lines added                     \\
                                                                             & deletions                  & Lines deleted                   \\
                                                                             & commits                     & Commit count                    \\ \hline
\multirow{6}{*}{\begin{tabular}[c]{@{}c@{}}Description\\ Style\end{tabular}} & char                  & Total characters                \\
                                                                             & header                & Markdown headers count          \\
                                                                             & list                  & List items count                \\
                                                                             & code                  & Code blocks count               \\
                                                                             & emoji                 & Emoji count                     \\
                                                                             & politeness               & Polite phrases count            \\ \hline
\begin{tabular}[c]{@{}c@{}}PR\\ Compliance\end{tabular}                     & conventional\_commit        & Conventional Commits compliance \\
\midrule
\end{tabular}%
}
\end{table}

Since the 11 extracted features vary in scale and unit (e.g., counts, densities, and binary values), we normalized them using Z-scores ($z = \frac{x - \mu}{\sigma}$), where $\mu$ and $\sigma$ denote the mean and standard deviation of the entire dataset, to facilitate direct comparison.
We then calculated the mean Z-score of each feature to analyze trends and relative characteristics across agents, visualizing the results in a heatmap.

\subsection{Human Reviewer Response (RQ2)}

To analyze the collaboration between human reviewers and AI coding agents, we divided RQ2 into two sub-questions.

\smallskip\noindent\textbf{RQ2-1.}
To analyze the nature of reviewer engagement—specifically engagement levels, feedback depth, and sentiment—we retrieved a total of 94,865 records from the AIDev dataset. This dataset comprised 39,122 entries from \textit{pr\_comments}, 28,875 from \textit{pr\_reviews}, and 26,868 from \textit{pr\_review\_comments\_v2}.

To ensure the analysis accurately reflects human responses, we performed a rigorous filtering process. We excluded 52,123 records associated with bot accounts and removed 13,781 records containing null values or consisting solely of whitespace. Consequently, the final dataset for the analysis consists of 28,961 records.

We defined four primary metrics to capture the nuances of interaction. Reviewer engagement was assessed using \textit{comments\_per\_pr}, the mean number of comments calculated across all 33,596 PRs. The responsiveness of reviewers was measured by \textit{time\_to\_first\_comment}, the median time elapsed from PR creation to the first reviewer comment for the 7,408 PRs that received feedback. To approximate the depth of feedback, we calculated \textit{comment\_length} based on the median character count of the 28,961 retained comments.

To evaluate the tone of the feedback, we employed a RoBERTa model fine-tuned on software engineering texts~\cite{zhang2020sentiment,novielli2020dataset}. This model classifies comments into \textit{positive}, \textit{neutral}, or \textit{negative} categories. The sentiment reflects linguistic tone rather than technical correctness.
For this specific analysis, we restricted the dataset to English-language comments to maintain classification validity. The language filtering process removed 1,573 records from \textit{pr\_comments}, 1,014 from \textit{pr\_reviews}, and 1,561 from \textit{pr\_review\_comments\_v2}, resulting in a subset of 24,813 records. Table~\ref{tab:agent_comments_rq2} shows the distribution of these comments across the five agents.

\begin{table}[tb]
\centering
\caption{Comment distribution across AI coding agents}
\label{tab:agent_comments_rq2}
\resizebox{0.95\columnwidth}{!}{%
\begin{tabular}{lrr}
\toprule
Agent & for Engagement Metrics & for Sentiment Analysis \\
\midrule
Claude Code & 902 & 790 \\
Github Copilot & 16,581 & 14,515 \\
Cursor & 1,254 & 1,083 \\
Devin & 6,603 & 5,536 \\
OpenAI Codex & 3,621 & 2,889 \\
\midrule
Total & 28,961 & 24,813 \\
\bottomrule
\end{tabular}%
}
\end{table}

\smallskip\noindent\textbf{RQ2-2.}
While RQ2-1 focuses on the process of review, RQ2-2 examines the final results of the PRs. We analyzed the complete set of 33,596 PRs to determine success rates and processing efficiency.

We employed two key metrics for this comparison. The \textit{merge\_\allowbreak rate} metric quantifies the percentage of PRs that were successfully merged across the dataset. To assess the efficiency of the review cycle, we measured \textit{time\_to\_completion}. This metric represents the median duration from PR creation to its final state for the 31,284 PRs that were either closed or merged. These metrics collectively provide insight into the acceptance and processing speed of code contributions generated by different AI coding agents.

\section{Results}
\subsection{PR Description Characteristics (RQ1)}

Claude Code, GitHub Copilot, Cursor, and Devin share a common characteristic: they generate descriptions with limited use of \textit{headers} and \textit{lists}, which is associated with reduced readability.

Claude Code and GitHub Copilot exhibit similar traits, both prioritizing code generation volume with substantial code modifications. Specifically, Claude Code strictly adheres to PR regulations. While its descriptions contain a significantly high volume of text, it appears to leverage \textit{emojis} to enhance readability. GitHub Copilot similarly produces a high volume of text but incorporates frequent \textit{code} blocks, suggesting a tendency to explicitly convey code changes and highlight critical sections to reviewers.

Cursor and Devin also share similar behavioral tendencies. As their code modification volumes are average, they do not appear to prioritize high-volume code generation. Cursor is likely to generate descriptions consisting primarily of plain text; however, it exhibits high \textit{politeness}, indicating a style that prioritizes a courteous tone towards developers. Devin strictly adheres to PR regulations but creates frequent commit splits relative to the volume of code changes, which may necessitate that reviewers examine numerous commits.

In contrast, OpenAI Codex is the only agent that frequently utilizes \textit{headers} and \textit{lists}, likely generating structured and highly understandable description characteristics. Furthermore, its lower volume of code changes and minimal commit splitting suggest a high probability of a streamlined review process.

\begin{figure}[tb]
\centering
\includegraphics[width=\columnwidth]{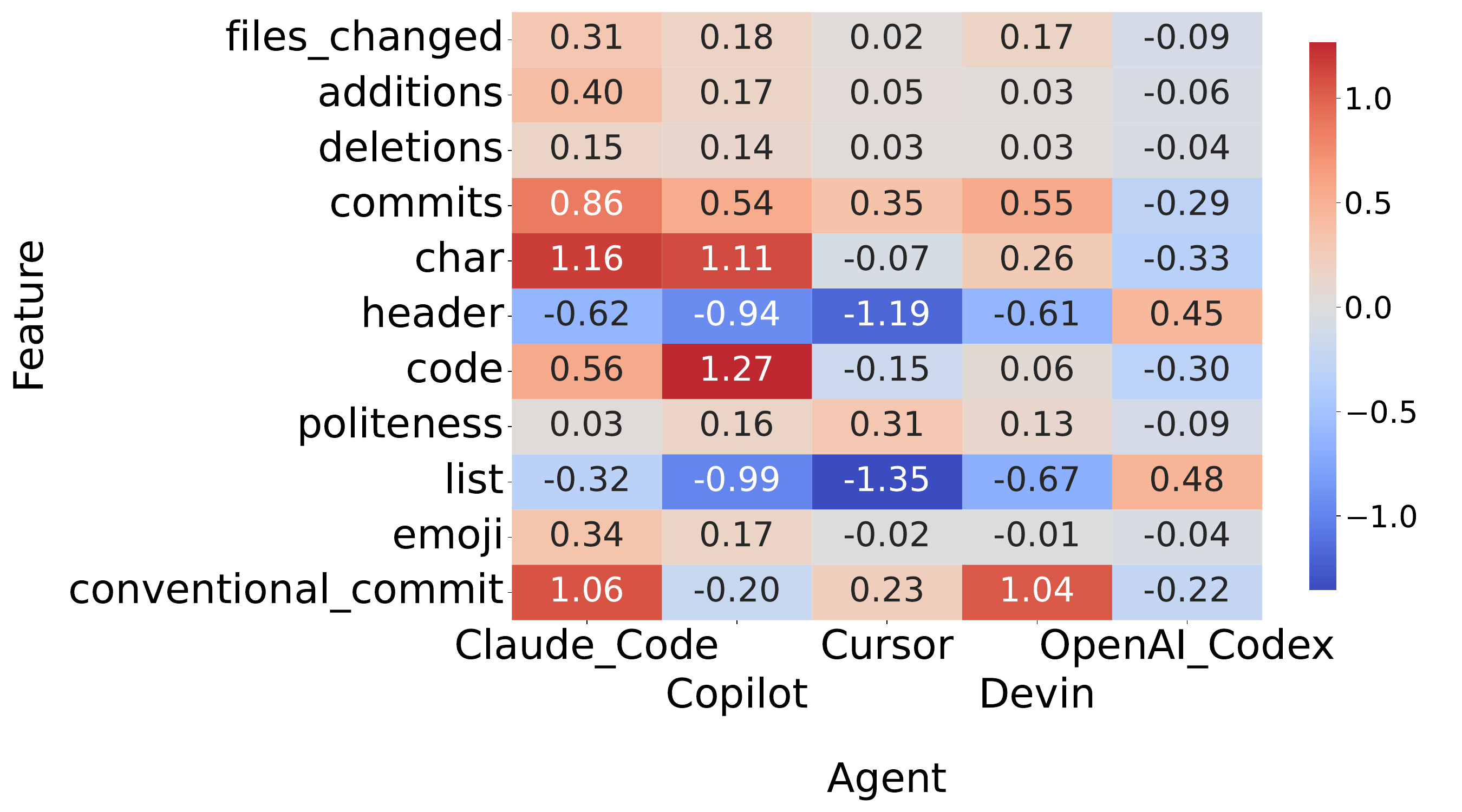}
\caption{Differences Among AI coding agents}
\label{fig:RQ1results}
\end{figure}

\begin{tcolorbox}[colback=gray!5,colframe=awesome,title=RQ1 Summary]

Answering \textbf{RQ1}, our analysis of PR description characteristics using 11 distinct features revealed a diverse spectrum of styles, ranging from code-centric, information-dense approaches to those prioritizing politeness and structural adherence. This variability suggests that AI coding agents possess distinct profiles regarding information volume, readability, and interaction style.
\end{tcolorbox}

\subsection{Human Reviewer Response (RQ2)}

\begin{figure}[tb]
\centering
\includegraphics[width=\columnwidth]{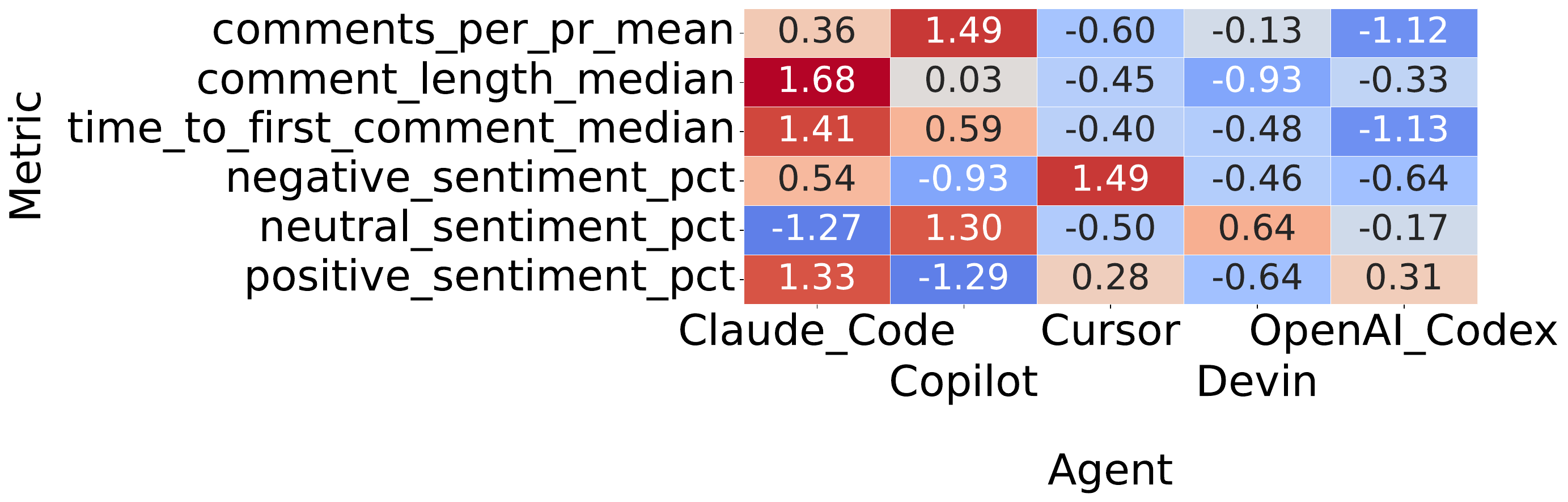}
\caption{Human reviewer interaction patterns}
\label{fig:RQ2results}
\end{figure}

\smallskip\noindent\textbf{RQ2-1.}
Figure~\ref{fig:RQ2results} visualizes standardized Z-scores for the four interaction metrics across the five agents. The sentiment metric is displayed separately for positive, neutral, and negative categories.

Reviewer engagement patterns varied across agents. Claude Code and GitHub Copilot both generated high levels of reviewer activity, though with different characteristics. Claude Code elicited the longest comments and the highest proportion of positive sentiment, while GitHub Copilot received the most comments per PR with predominantly neutral sentiment. In contrast, Devin and OpenAI Codex both received minimal engagement, with short response times and either brief comments or few comments overall. Cursor stood apart as the agent receiving the highest proportion of negative sentiment.

Table~\ref{tab:statistical_tests} summarizes the statistical test results for the RQ2 metrics.
Regarding the interaction process (RQ2-1), we observed significant differences across agents for all metrics ($p < 0.001$).
Notably, \textit{comments\_per\_pr} exhibited a medium-to-large effect size ($\varepsilon^2 = 0.280$), whereas \textit{comment\_length} and \textit{time\_to\_first\_comment} showed small or negligible effect sizes ($\varepsilon^2 = 0.006$ and $0.018$, respectively).

\begin{table}[tb]
\centering
\caption{Statistical test results for RQ2 metrics}
\label{tab:statistical_tests}
\resizebox{0.95\columnwidth}{!}{%
\begin{tabular}{llclr}
\toprule
Metric & Test & $p$-value & Effect Size & $n$ \\
\midrule
\multicolumn{5}{l}{\textit{\textbf{RQ2-1: Interaction Process}}} \\
Comments / PR & KW & $<0.001$ & $\varepsilon^2 = 0.280$ & 33,596 \\
Comment Length & KW & $<0.001$ & $\varepsilon^2 = 0.006$ & 28,961 \\
Time to 1st Comment & KW & $<0.001$ & $\varepsilon^2 = 0.018$ & 7,408 \\
Sentiment Dist. & $\chi^2$ & $<0.001$ & $V = 0.128$ & 24,813 \\
\midrule
\multicolumn{5}{l}{\textit{\textbf{RQ2-2: Outcome}}} \\
Merge Rate & $\chi^2$ & $<0.001$ & $V = 0.349$ & 33,596 \\
Time to Completion & KW & $<0.001$ & $\varepsilon^2 = 0.325$ & 31,284 \\
\bottomrule
\multicolumn{5}{r}{\footnotesize KW: Kruskal-Wallis test, $\chi^2$: Chi-square test} \\
\end{tabular}%
}
\end{table}

\smallskip\noindent\textbf{RQ2-2.}
Table~\ref{tab:outcome_metrics_rq2_2} shows the merge rate and completion median time for each agent. Table~\ref{tab:statistical_tests} presents the statistical test results for RQ2-2. We observed significant differences across agents with medium-to-large effect sizes for both \textit{merge\_rate} and \textit{time\_to\_completion}.

Outcome metrics revealed two distinct patterns. OpenAI Codex and Cursor both achieved favorable results. Codex showed the highest merge rate and shortest completion time, while Cursor maintained the second-highest merge rate despite receiving negative sentiment. In contrast, GitHub Copilot and Devin both showed unfavorable outcomes with lower merge rates and extended completion times. Copilot generated extensive discussion yet had the lowest merge rate, suggesting inefficient review cycles, while Devin received minimal engagement and tended to be closed without resolution. Claude Code elicited in-depth discussions but reached a resolution quickly.

\begin{table}[tb]
\centering
\caption{Outcome metrics by agent for RQ2-2}
\label{tab:outcome_metrics_rq2_2}
\resizebox{0.95\columnwidth}{!}{%
\begin{tabular}{lrr}
\toprule
Agent & Merge Rate (\%) & Time to Completion (hours) \\
\midrule
Claude Code & 59.0 & 1.95  \\
GitHub Copilot & 43.0 & 13.0  \\
Cursor & 65.22 & 0.90  \\
Devin & 53.76 & 8.91  \\
OpenAI Codex & 82.6 & 0.02  \\
\bottomrule
\end{tabular}
}
\end{table}

\begin{tcolorbox}[colback=gray!5,colframe=awesome,title=RQ2 Summary]
Answering \textbf{RQ2}, our analysis of human reviewer response patterns and outcome metrics revealed statistically significant differences across agents regarding engagement levels, feedback depth, and sentiment (RQ2-1), as well as merge rates and time to completion (RQ2-2).
These findings suggest that the choice of AI agent influences both the human reviewer's workload and the final PR outcomes.
\end{tcolorbox}

\section{Threats to Validity}\label{sec:threats_to_validity}

\noindent\textbf{Threats to internal validity: }
Our analysis is observational and does not support causal claims. Differences across agents may be confounded by task type, repository norms, code complexity, and RQ1 is not conditioned on task categories.

\noindent\textbf{Threats to construct validity: }
We approximate PR description characteristics via automated parsing which may imperfectly capture description details. Additionally, filtering non-English comments for sentiment analysis may introduce bias in sentiment distributions across agents.

\noindent\textbf{Threats to external validity: }
Results are drawn from AIDev.
Generalization to other agents, time periods, and non GitHub workflows may be limited.

\section{Related Work}\label{sec:relatedwork}

Studies analyze developer response patterns in development environments before AI coding agents arrived. In human PRs, research shows a strong correlation between the time until a reviewer posts the first comment and the time until the PR finally closes\cite{yue2015}. Toxic interactions, including insults, reduce productivity and sustainability, and they hinder the efficiency of the development process\cite{jaydeb2025}. Patterns show that bug insertion rates increase when emotional language dominates the discussion. This suggests that neutral communication that suppresses emotional fluctuations plays an important role in maintaining development efficiency and quality\cite{syed2019}.In code reviews, the emotions (positive or negative) that developer comments contain correlate with the time until review completion and code merging\cite{ikram2019}.

Prior research has analyzed developer responses before the introduction of AI coding agents, but researchers have not yet analyzed responses in PRs incorporating AI coding agents. Thus, this study focuses on human responses in agent-assisted development to derive insights for enhancing development efficiency in the AI era.

\section{Discussion and Conclusion}
\noindent\textbf{Discussion.}
Our results show a statistical association between PR structure and review efficiency, with more structured PRs associated with faster reviewer responses and completion times. This suggests that clearer organization may help reviewers more quickly understand the intent of changes. These associations may also reflect simpler changes or higher-quality code rather than description structure alone. However, PR readability alone does not determine merge outcomes. We observed PRs with limited structure and negative sentiment that were still merged, indicating that code quality remains a key factor in acceptance. We observe statistical associations between PR description characteristics and reviewer responses.

Reviewer sentiment should be interpreted with caution. Negative comments often target presentation issues rather than rejecting the code itself, while neutral comments on unmerged PRs may reflect limited reviewer engagement rather than approval.
Accordingly, sentiment is contextual and should not be treated as a proxy for PR acceptance or quality.
Finally, our findings are observational and may be influenced by confounding factors such as repository norms, task complexity, or deployment contexts of AI coding agents; we therefore describe descriptive patterns of human–AI interaction rather than causal effects.

\noindent\textbf{Conclusion.} 
This study shows that, in AI-generated pull requests, how changes are communicated is associated with the review process, in addition to the functional correctness of the code. More structured PR descriptions are associated with faster reviewer responses and shorter completion times, suggesting that clear organization can reduce reviewer effort. However, presentation alone does not determine acceptance: structured descriptions do not guarantee merges, and code quality remains a central factor in PR outcomes. Overall, these findings highlight the role of effective communication in human–AI collaborative software development while emphasizing that successful integration ultimately depends on both clear explanations and high-quality code.


\begin{acks}
This work was supported by JSPS KAKENHI No. JP24K14895 and Support Center for Advanced Telecommunications Technology Research.
\end{acks}

\bibliographystyle{ACM-Reference-Format}
\bibliography{reference}

@article{li2025aidev,
  title={The Rise of AI Teammates in Software Engineering (SE) 3.0: How Autonomous Coding Agents Are Reshaping Software Engineering},
  author={Li, Hao and Zhang, Haoxiang and Hassan, Ahmed E.},
  journal={arXiv preprint arXiv:2507.15003},
  year={2025}
}

@misc{github_copilot,
  author       = {{GitHub}},
  title        = {{GitHub Copilot}},
  howpublished = {\url{https://github.com/features/copilot}},
  year         = {2026},
  note         = {Accessed: 2026-01-24}
}

@misc{cursor,
  title = {Cursor},
  author = {{Anysphere}},
  year = {2025},
  url = {https://cursor.com/},
  note = {Accessed: 2025-12-17}
}

@article{DBLP:journals/corr/abs-2110-15447,
  author       = {SayedHassan Khatoonabadi and
                  Diego Elias Costa and
                  Rabe Abdalkareem and
                  Emad Shihab},
  title        = {On Wasted Contributions: Understanding the Dynamics of Contributor-Abandoned
                  Pull Requests},
  journal      = {CoRR},
  volume       = {abs/2110.15447},
  year         = {2021},
  url          = {https://arxiv.org/abs/2110.15447},
  eprinttype    = {arXiv},
  eprint       = {2110.15447},
  bibsource    = {dblp computer science bibliography, https://dblp.org}
}

@inproceedings{zhang2020sentiment,
  author={Zhang, Ting and Xu, Bowen and Thung, Ferdian and Haryono, Stefanus Agus and Lo, David and Jiang, Lingxiao},
  booktitle={2020 IEEE International Conference on Software Maintenance and Evolution (ICSME)}, 
  title={Sentiment Analysis for Software Engineering: How Far Can Pre-trained Transformer Models Go?}, 
  year={2020},
  volume={},
  number={},
  pages={70-80},
  doi={10.1109/ICSME46990.2020.00017}
}

@inproceedings{novielli2020dataset,
author = {Novielli, Nicole and Calefato, Fabio and Dongiovanni, Davide and Girardi, Daniela and Lanubile, Filippo},
title = {Can We Use SE-specific Sentiment Analysis Tools in a Cross-Platform Setting?},
year = {2020},
isbn = {9781450375177},
publisher = {Association for Computing Machinery},
address = {New York, NY, USA},
url = {https://doi.org/10.1145/3379597.3387446},
doi = {10.1145/3379597.3387446},
booktitle = {Proceedings of the 17th International Conference on Mining Software Repositories},
pages = {158–168},
numpages = {11},
location = {Seoul, Republic of Korea},
series = {MSR '20}
}

@misc{hassan2024ainativesoftwareengineeringse,
      title={Towards AI-Native Software Engineering (SE 3.0): A Vision and a Challenge Roadmap}, 
      author={Ahmed E. Hassan and Gustavo A. Oliva and Dayi Lin and Boyuan Chen and Zhen Ming and Jiang},
      year={2024},
      eprint={2410.06107},
      archivePrefix={arXiv},
      primaryClass={cs.SE},
      url={https://arxiv.org/abs/2410.06107}, 
}

@article{DBLP:journals/corr/abs-2504-18407,
  author       = {Yoseph Berhanu Alebachew and
                  Minhyuk Ko and
                  Chris Brown},
  title        = {Are We on the Same Page? Examining Developer Perception Alignment
                  in Open Source Code Reviews},
  journal      = {CoRR},
  volume       = {abs/2504.18407},
  year         = {2025}
}

@inproceedings{yang2024sweagentagentcomputerinterfacesenable,
      title={SWE-agent: Agent-Computer Interfaces Enable Automated Software Engineering},
      author={Yang, John and Jimenez, Carlos E. and Wettig, Alexander and Lieret, Kilian and Yao, Shunyu and Narasimhan, Karthik and Press, Ofir},
      booktitle={Advances in Neural Information Processing Systems 37 (NeurIPS 2024)},
      year={2024},
      publisher={Curran Associates, Inc.},
      doi={10.52202/079017-1601},
      url={https://doi.org/10.52202/079017-1601}
}

@inproceedings{Mozannar2024,
author = {Mozannar, Hussein and Bansal, Gagan and Fourney, Adam and Horvitz, Eric},
title = {Reading Between the Lines: Modeling User Behavior and Costs in AI-Assisted Programming},
year = {2024},
isbn = {9798400703300},
publisher = {Association for Computing Machinery},
address = {New York, NY, USA},
url = {https://doi.org/10.1145/3613904.3641936},
doi = {10.1145/3613904.3641936},
booktitle = {Proceedings of the 2024 CHI Conference on Human Factors in Computing Systems},
articleno = {142},
numpages = {16},
location = {Honolulu, HI, USA},
series = {CHI '24}
}

@misc{peng2023impactaideveloperproductivity,
      title={The Impact of AI on Developer Productivity: Evidence from GitHub Copilot}, 
      author={Sida Peng and Eirini Kalliamvakou and Peter Cihon and Mert Demirer},
      year={2023},
      eprint={2302.06590},
      archivePrefix={arXiv},
      primaryClass={cs.SE},
      url={https://arxiv.org/abs/2302.06590}, 
}

@misc{shah2025evolutionprogrammerstrustgenerative,
      title={Evolution of Programmers' Trust in Generative AI Programming Assistants}, 
      author={Anshul Shah and Thomas Rexin and Elena Tomson and Leo Porter and William G. Griswold and Adalbert Gerald Soosai Raj},
      year={2025},
      eprint={2509.13253},
      archivePrefix={arXiv},
      primaryClass={cs.HC},
      url={https://arxiv.org/abs/2509.13253}, 
}

@misc{tufano2024autodevautomatedaidrivendevelopment,
      title={AutoDev: Automated AI-Driven Development}, 
      author={Michele Tufano and Anisha Agarwal and Jinu Jang and Roshanak Zilouchian Moghaddam and Neel Sundaresan},
      year={2024},
      eprint={2403.08299},
      archivePrefix={arXiv},
      primaryClass={cs.SE},
      url={https://arxiv.org/abs/2403.08299}, 
}

@misc{he2025doesaiassistedcodingdeliver,
      title={Does AI-Assisted Coding Deliver? A Difference-in-Differences Study of Cursor's Impact on Software Projects}, 
      author={Hao He and Courtney Miller and Shyam Agarwal and Christian Kästner and Bogdan Vasilescu},
      year={2025},
      eprint={2511.04427},
      archivePrefix={arXiv},
      primaryClass={cs.SE},
      url={https://arxiv.org/abs/2511.04427}, 
}

@inproceedings{cihan2024automatedcodereviewpractice,
  title={Automated Code Review In Practice},
  author={Cihan, Umut and Haratian, Vahid and I{\c{c}}{\"o}z, Arda and G{\"u}l, Mert Kaan and Devran, {\"O}mercan and Bayendur, Emircan Furkan and U{\c{c}}ar, Baykal Mehmet and T{\"u}z{\"u}n, Eray},
  booktitle={2025 IEEE/ACM 47th International Conference on Software Engineering: Software Engineering in Practice (ICSE-SEIP)},
  year={2025},
  doi={10.1109/ICSE-SEIP66354.2025.00043},
  url={https://doi.org/10.1109/ICSE-SEIP66354.2025.00043},
  publisher={IEEE}
}

@misc{bakal2025experiencegithubcopilotdeveloper,
      title={Experience with GitHub Copilot for Developer Productivity at Zoominfo}, 
      author={Gal Bakal and Ali Dasdan and Yaniv Katz and Michael Kaufman and Guy Levin},
      year={2025},
      eprint={2501.13282},
      archivePrefix={arXiv},
      primaryClass={cs.SE},
      url={https://arxiv.org/abs/2501.13282}, 
}

@misc{vaillant2024developersperceptionsimpactchatgpt,
      title={Developers' Perceptions on the Impact of ChatGPT in Software Development: A Survey}, 
      author={Thiago S. Vaillant and Felipe Deveza de Almeida and Paulo Anselmo M. S. Neto and Cuiyun Gao and Jan Bosch and Eduardo Santana de Almeida},
      year={2024},
      eprint={2405.12195},
      archivePrefix={arXiv},
      primaryClass={cs.SE},
      url={https://arxiv.org/abs/2405.12195}, 
}

@inproceedings{syed2019,
author = {Huq, Syed Fatiul and Sadiq, Ali and Sakib, Kazi},
year = {2019},
month = {12},
pages = {514-521},
title = {Understanding the Effect of Developer Sentiment on Fix-Inducing Changes: An Exploratory Study on GitHub Pull Requests},
doi = {10.1109/APSEC48747.2019.00075}
}

@INPROCEEDINGS{yue2015,
  author={Yu, Yue and Wang, Huaimin and Filkov, Vladimir and Devanbu, Premkumar and Vasilescu, Bogdan},
  booktitle={2015 IEEE/ACM 12th Working Conference on Mining Software Repositories}, 
  title={Wait for It: Determinants of Pull Request Evaluation Latency on GitHub}, 
  year={2015},
  volume={},
  number={},
  pages={367-371},
  doi={10.1109/MSR.2015.42}}

@misc{jaydeb2025,
author = {Sarker, Jaydeb and Turzo, Asif Kamal and Bosu, Amiangshu},
year = {2025},
month = {02},
pages = {},
title = {The Landscape of Toxicity: An Empirical Investigation of Toxicity on GitHub},
doi = {10.48550/arXiv.2502.08238}
}

@article{ikram2019,
author = {Asri, Ikram El and Kerzazi, Noureddine and Uddin, Gias and Khomh, Foutse and Janati Idrissi, M.A.},
title = {An empirical study of sentiments in code reviews},
year = {2019},
issue_date = {Oct 2019},
publisher = {Butterworth-Heinemann},
address = {USA},
volume = {114},
number = {C},
issn = {0950-5849},
url = {https://doi.org/10.1016/j.infsof.2019.06.005},
doi = {10.1016/j.infsof.2019.06.005},
journal = {Inf. Softw. Technol.},
month = oct,
pages = {37–54},
numpages = {18}
}

@article{barke2023grounded,
  title={Grounded copilot: How programmers interact with code-generating models},
  author={Barke, Shraddha and James, Michael B and Polikarpova, Nadia},
  journal={Proceedings of the ACM on Programming Languages},
  volume={7},
  number={OOPSLA1},
  pages={85--111},
  year={2023},
  publisher={ACM New York, NY, USA}
}

\end{document}